# Developing a model for a text database indexed pedagogically for teaching the Arabic language

Asma Boudhief(1), Mohssen Maraoui(2), Mounir Zrigui(3)

(1) UTIC laboratory, Higher School of Sciences and Techniques of Tunis, Tunisia
asmaboudhief@live.com
(2) UTIC Laboratory, University of Monastir, Tunisia
maraoui.mohsen@gmail.com
(3) UTIC Laboratory, University of Monastir, Tunisia
mounir.zrigui@fsm.rnu.tn

**Abstract.** In this memory we made the design of an indexing model for Arabic language and adapting standards for describing learning resources used (the LOM and their application profiles) with learning conditions such as levels education of students, their levels of understanding… the pedagogical context with taking into account the representative elements of the text, text's length,... in particular, we highlight the specificity of the Arabic language which is a complex language, characterized by its flexion, its voyellation and its agglutination.

## 1 Introduction

With the advancing technology, the teaching of the language has undergone great changes and teachers use computers to better present their course, while the existing systems for the Arabic language does not meet their needs, there are static systems, characterized by the absence of auto-correction and the absence of changes in exercises in the same unit of learning.

Presenting a course that meets the needs of teachers requires the right choice of text, which seems difficult because of the lack of tools allowing access to the texts according to desired criteria.

"Although the text search seems to be a recurring tasks in language teaching, it seems that few tools have been designed to enable teachers to access texts based on criteria related to their problems "(Loiseau et al., 2008) [1].

Exists Mechanisms of search are based on a traditional search by keywords and this mechanism seems inefficient, it requires a pedagogic text's indexing to facilitate his search.

# 2 State of the Art

## 2.1 History of ALAO

The ALAO (Learning of Language Assisted by Computer) is a field of research and development that interest to several disciplines involved in the field of cognitive science:

- linguistics (in its broadest sense including theoretical linguistics America, Applied Language Teaching)
- the computational linguistics
- Informatique Technology (especially artificial intelligence (AI))
- the psycholinguistique.[2]

What is often considered the first teaching machine is that developed by Sidney Pressey in the 20s. It's just an automated machine to correct multiple choices, with four buttons (one per response). When the student press the right button it switches to the next question.[3]

But the starting point of EIAH (environments for human learning) is the programmed learning.

❖ The programmed learning: is "a teaching method" that transmits knowledge directly without the intermediary of a teacher or a monitor, this while respecting the characteristics of individual students. This teaching is based on four principles:
- The principle of structuring the subject to be taught (this is to cut and present the subject so as to facilitate comprehension and memorization).
- The principle of adaptation (teaching must be adapted to the student)
- The principle of stimulation
- The principle of control

❖ Computer Assisted Education (French abbreviation: EAO) is actually born in the early 60s and it is only in the 70s, the work on expert systems appear the first attempts to make "intelligent "EAO. This research aims was to bridge the existing limits. Then the 70's were marked by the first micro-monde, the 80 by intelligent tutors and 90 years by cooperative systems and interactive learning environments with computer.[4]

## 2.2 Contribution of TAL to ALAO

The potential contribution of TAL[5] comes from its issue and the goal he has set. Uniquely it allows considering the linguistic form, not as a sequence of signs devoid of interpretation, but as element of system with two levels (form and meaning). A system of lan-guage learning cannot be valid or acceptable if it is able, first to generate only correct linguistic knowledge (not false) and secondly to appraise correctly the language productions of learners.

TAL:

- ➢ May provide tools for both the evaluation and the automatic creation of activities.
- ➢ Allow system to be able to detect, explain and automatically correct errors of the learner if we want it to work independently

## 2.3 Model of information system

*"An IRS (Information Retrieval System) is a computerized system that facilitates access to a set of documents (corpus), to help find those whose content best fits for need of information of a user."* [6]

In an IRS (see Figure 1) we perform in the first hand, an indexing of existing documents in the database to obtain a model of documents. On the other hand if a user sends a request, it will be interpreted and the system creates facets representing this query. Then it performs a match with the model of documents to extract the most relevant documents at the request of the user.

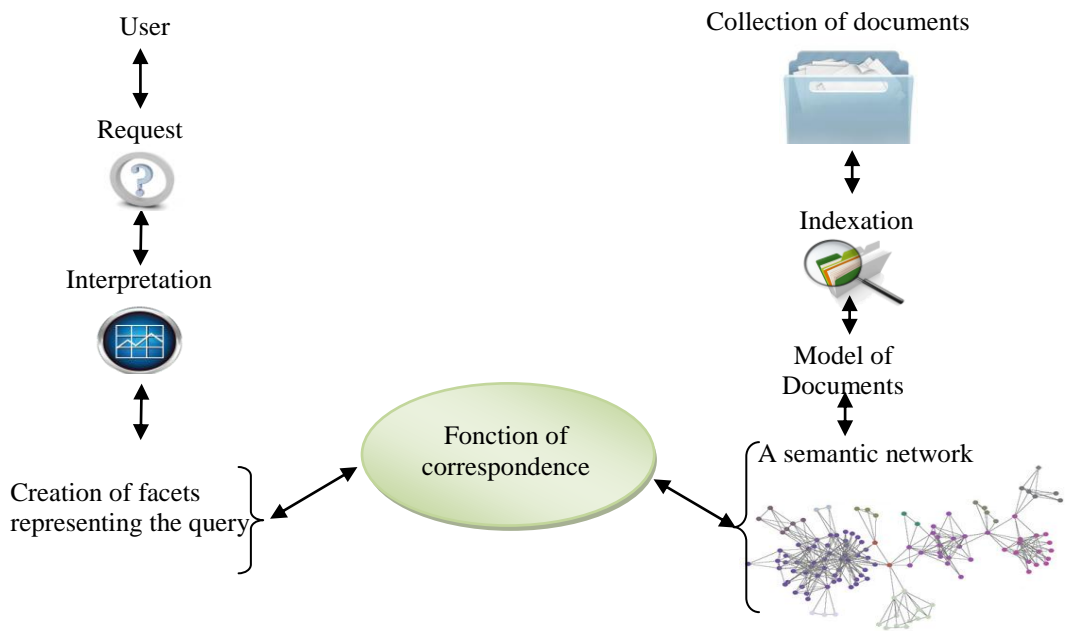

**Figure1:** General scheme of a model of information retrieval

For each system, you must:
- ➢ Define what type of documents will be processed. i. e: paragraph, text page or multiple pages.
- ➢ Establish an indexing by extracting words deemed relevant in defining the document in order to obtain templates.
- ➢ Creation of semantic networks to present the document templates

On the other hand, a request coming from the user should be construed to create representative facets, and then the system establishes the correspondence between the created facets and models of the semantic network by calculating the degree of similarity[7] and releasing the document nearest to the user's request.

## 2.4 pedagogical indexing

According to Loiseau, pedagogical indexing[8] is an "indexing performed according to a documentary language that allows the user to search for objects to use in education".[9]
Our thesis aims to propose a model for what we called the indexation of pedagogical texts for teaching the Arabic language[10] and demonstrate its feasibility by implementing a prototype.

This leads us to insist that such a database must allow the following use cases:
- ➢ adding text to the base
- ➢ Text searching based on the problematic issues and specificity of the Arabic language.
- ➢ support for selecting text

The indexing operation which consists of analyzing the object to be indexed by "extracting concepts" of this analysis, and finally express it in a documentary language.
The agent of these operations is not specified. So we can imagine several configurations: the analysis can be performed by a human operator or a machine, and the expression of concepts extracted in documentary language can be performed either by a human or a machine.[9]

In our case, the indexing operation will involve the user and the system. The user is not a documentalist and as a teacher, it is primarily the use case "text's search" which interests him. Both sides of the indexing process will therefore be as simple and not boring as possible and to do that, they must be automated as possible.

The analysis of some concepts of the document cannot be automated, such as the author or title (if these criteria are relevant to their operation in language teaching). But any automated analysis must be supported by the system. The analysis part of the indexing will be hybrid in the sense that some concepts cannot be managed by the system, but the most fastidious will be automated where possible.

In what follows, we will explain the influence of pedagogical context on the choice of the text, before exploring the existing standard description of educational resources so that we may adapted to the specificity of the Arabic language and needs users, and then introduce the notion of facet of a text and present a model in which it occurs.

### 2.5 Influence of pedagogic context in the choice of the text

After the formulation of the problem (setting up an activity), a text was assigned successively projected properties. We call these properties specified at progressively steps, the learning environment. The pedagogical context is defined as "the set of features describing the teaching situation".[11]

As part of the educational indexing, it is not appropriate to set all properties that may intervene in the educational context (CP), but rather to try to identify the relevant components of the text search as shown in Figure 2.

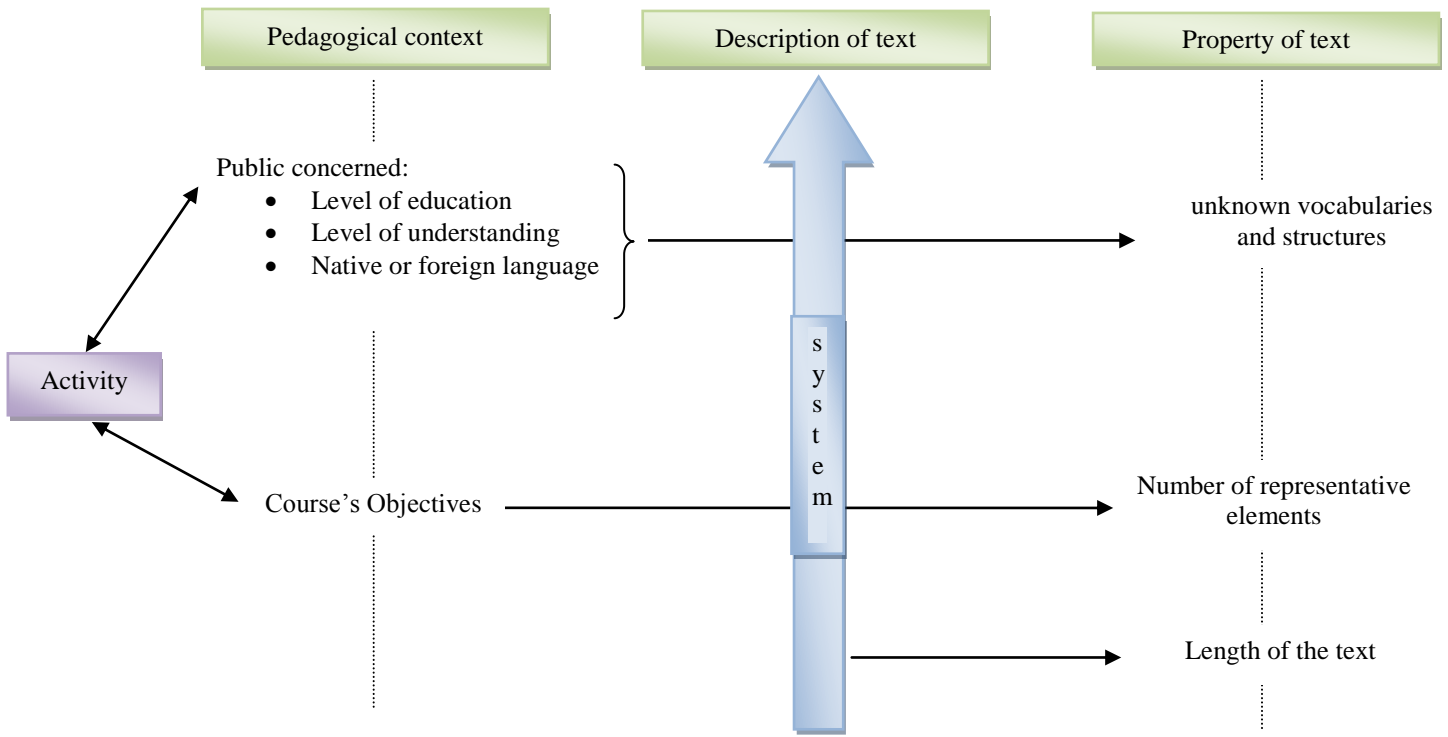

**Figure2**: Effect of pedagogical context on assigning properties to the text

According to the diagram, we see that the assignment of properties to the text depends to:
- The educational context: he relates, firstly the main objectives of the course (during conjugation of verbs, grammar courses ...) and secondly the targets publics as the level of students, their understanding and the studied language (native or foreign).
- The description of the text: which we can determine the length of the text which is an important criterion of language education, and its title.

From these two criteria, we can determine some properties of text such as:
- Its length: the majority of language teachers want to have a significant text and minimum length to not annoy the learner.
- The representative elements: for example if the student requests an exercise for the conjugation of an Arabic verb composed of three letters in the past, our representative element is the verb.
- The unknown vocabularies for students: for example if the student knows the verbs composed of three letters in Arabic but he does not know the conjugation of its verbs "mootalla" they contain the letter "alif" (ا), so we must consider this case.

And then make the decision to choose the text in a specific context.

## 2.6 The standard description of educational resources

The indexing of learning objects is an absolute necessity if we want to find them. For that, you have to add semantic information. This information is metadata: data describing data[12].

For that metadata fulfill their role and facilitate access to online resources, it is imperative that a stable standard that exists to providers of resource and users can use the same repository. This standard must also support developments and extensions to accommodate new needs.

*The Learning Object Metadata* (LOM) is a standard published in 2002 by *the Learning Technology Standards Commitee* (LTSC) de l'IEEE (*Institute of Electrical and Electronics Engineers*). The standard consists of four parts:
• IEEE 1484.12.1 - Conceptual model of metadata;
• IEEE 1484.12.2 - Implementation of ISO / IEC 11404 in the LOM metadata model;
• IEEE 1484.12.3 - Development and implementation of the XML Schema for LOM;
• IEEE 1484.12.4 - Definition of application framework RDF (Resource Description Framework) for LOM.[13]

All LOM elements are optional, that is to say that the model can work without all the fields are filled. Nevertheless, it is desirable to provide the most information so that resources can be exploited.

The LOM is organized into nine categories performing different functions. The elements contained in each category can be seen in Figure 3.

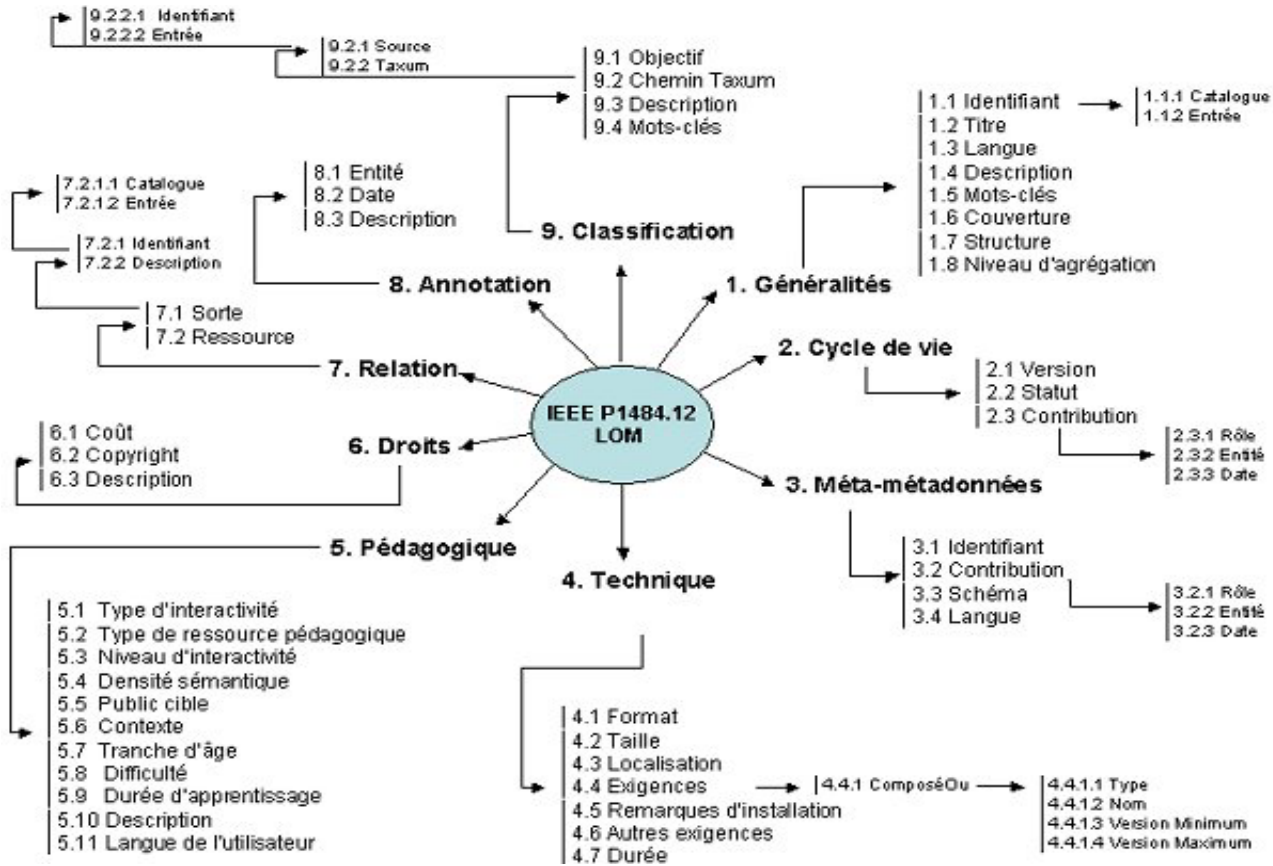

**Figure3:** Organization of LOM metadata scheme.[14]

The LOM standard is:
- Fairly complex because it includes 78 elements forming a tree on three floors and offers 59 fields of information independent.
- An Abstract model which must be instantiated in a particular context. It is therefore to adapt this standard to respond to specific and real needs of users.

This work of adaptation led to the definition of an "application profile".

*"An application profile is a set of elements selected from one or more metadata schemas and combined in a composite diagram ... Its purpose is to adapt existing schemes to provide a tailor-made to all the functional requirements of a particular application, while remaining interoperable with the original patterns "*[15]

All existing application profiles have some common characteristics:
- ❖ Assigning a degree of obligation to schema elements (mandatory, recommended, optional)
- ❖ The addition of new elements characterizing the environment of use of resources.
- ❖ Defining vocabularies most appropriate to learning conditions (level of students, level of understanding ...).

Among the application profiles we quote:
- ➢ Normetic[16]: an application profile which terms of documentation elements include four statuses: required, conditionally required, recommended or optional. One of these four statutes was assigned to each element selected for Normetic.

- ➢ Celebrate[17]: There are two types of subsets of elements defined here: the elements that must be met in all instances of the metadata (mandatory elements) and elements that would be very useful to be filled (Recommended Elements) . All other elements of the set CELEBRATE are considered optional.

- ➢ SCORM[18]: Earlier versions of SCORM defined notion of a field mandatory or optional, while SCORM ® 2004 defined all elements of the data model as required. In other words, a consistent LMS SCORM ® 2004 must implement the entire data model.

- ➢ Cancore[19]: He only recommends the use of LOM elements for the purpose of interoperability in distributed environments.

## 3 Grammar, composition and characteristics of Arabic

Arabic is a language that:
- ➢ The words are written in horizontal lines from right to left.
- ➢ Most letters change shape depending on whether they appear at the beginning, middle or end of a word.
- ➢ The letters can be joined are always united both handwritten and printed.
- ➢ The vowels are added below and above the letters.

In Arabic, a sentence can be either nominal "إسمية" or verbal "فعلية".

### 3.1 Nominal phrase (الجملة الإسمية)

Consists of two parts: "al mobtada" and "al khabar"

#### 3.1.1 AL mobtada

In its simplest form is a name "mouaraf" and "marfouaa".
Example: المطرُ غزيرٌ (**the rain is pouring**). But it may take other simple or complex shapes.

  *a. Simple :*

- « ism alam » (noun): منير سعيد  '**Mounir is happy**'
- « Dhamir » (personal pronoun): أنتِ جادةٌ  '**you are serious**'
- « ichara ism » (demonstrative adjective): هذهِ فتاةٌ  '**it's a girl**'

  *b. Complex :*

- "mourakab naati" الأَشْجَارُ الْخَضْرَاءُ نَافِعَةٌ (**green trees are useful**)
- "mourakab idhafi" طَقْسُ الشِّتَاءِ بَارِد (**winter climate is cold**)
- "mourakab Badali" الْقَائِدُ حنّبعل يُحِبُّ وَطَنَهُ (**commander Hannibal loves his country**)
- "Mourakab atfi" أحمدُ و سَعِيدٌ لأعبَانِ (**Ahmed and Said are players**)
- "mourakab tawkidi" الأَوْلَادُ كُلُّهُمْ فِي سَاحَةِ الْمَدْرَسَةِ (**all children in the school yard**)
- "isem istefhem" أيُّ الموظفين انشطُ ؟ (**what is the staff member most active?**)
- "isem mawsoul" الذي جاءَ سامرٌ (**who arrived is samir**)
- "masdar" أنْ تقتصِدَ انفعُ لك (**saving is better for you**)

### 3.1.2 Al khabar

In its simplest form is a " mofrada." (single) **Example:** المطرُ غزيرٌ (**the rain is pouring**)
But it can be complex:
- "mourakeb jar ': القَمَرُ الصِّنَاعِيُّ فِي الْفَضَاءِ (**the satellite in space**)
- "mourakeb idhafi" الْعُصْفُورُ فَوْقَ الْغُصْنِ (**the bird on the branch**)
- "mourakeb atfi" الْقِصَّةُ قَصِيرَةٌ وَ مُمْتِعَةٌ (**the story is short and is fun**)
- "mourakeb isnedi" هَذَا يَجْرِي (**this court**)

Developing a model for a text database indexed pedagogically for teaching the Arabic language

## 3.2 Verbal phrase (الجملة الفعلية)

A verbal clause constitutes either of a verb + subject or a verb + subject + complement (object, time or place)

### 3.2.1 The verbe (الفعل)

In viewpoint tenses of conjugation, the verb in Arabic is composed of three types: past tense, future tense or order.
In view of the elements that follow, we can compose the verb in two types:
- "lezim" which requires only one subject
- "moutaadi" which requires a subject and an object

One can also compose the verbs in "Sahih" and "mouatal"
- "Sahih"   خرج (go out), عد (count), أخذ (take)
- "mouatal" which in its basic character is a letter of "illa" ("alif,"(ا) "ya"(ي) and "waw"(و))   يئس (despair), نام (sleep)

On the conjugation of verbs you can find a common structure between the verbs in the same category, taking into consideration the special cases such as the verbs "mahmouza" and the verbs "moatalla"....

**Example:** Verbs "thoulethi moujarrad" in Past and future

فَعَلَ- يَفْعُلُ : نَصَرَ - يَنْصُرُ
فَعَلَ- يَفْعِلُ : جَلَسَ - يَجْلِسُ
فَعَلَ- يَفْعَلُ: مَنَعَ - يَمْنَعُ
فَعِلَ- يَفْعَلُ: عَلِمَ - يَعْلَمُ
فَعِلَ- يَفْعِلُ: حَسِبَ - يَحْسَبُ
فَعُلَ- يَفْعُلُ: كَرُمَ – يَكْرُمُ

### 3.2.2 Subject (الفاعل)

It is a name "marfoua" that precedes the verb which means who release the action.
There may be several types:

- « Isem Dhaher moarab » طلَعَتِ الشَّمْسُ (the sun rises)
- « Isem Dhaher mabni »
  - « Dhamir » حضَرْت الاحتفالَ (she attended the festival)
  - « Isem ichara » سرّني هذا المشهد (I like this landscape)
  - « Isem mawsoul » غادر الذي رافقته (who you accompany left)
- « Dhamir mostatir » حضرْ نَفسكَ للاختبار (get ready (your self) for the test)

NB: as the case, the subject may precede « المفعول به » (object) as it can follow it.

### 3.2.3 Object complement (المفعول به)

It's over which it exercises the action, it is always "Mansoub" and can be explicit or implicit.

- ➢ Explicit:
  - "Isem Dhaher" كَرَّمَت الوزارة المعلمين (**Department honored the teachers**)
  - "Dhamir motassil" ساعدتُكَ في محنَتِكَ (**I helped you in your misfortune**)
  - "Dhamir monfassil" إياك نعبدُ (**to you that we do the prayers**)

- ➢ Implicit: عَرَفْتُ أنكَ قادم (**I know that you will come**)

### 3.2.4 Complement of time and place

- ➢ Complement of time; it is preceded by words such as: حين – سنة – ساعة – ظُهر – صباح
  إذا – متى– أيّانَ – إذْ – الآنَ – مذْ – منذُ – قطّ – بينما – ريثما – لمّا أمس
- ➢ Complement of place : it is preceded by words such as: فوق – تحت – أمام – وراء – حيث – دون...

### 3.2.5 Complement of manner (الحال)

It is a name "nakira" and "Mansoub" which expresses the subject's condition.
*Example:* سافر خالدٌ حزيناً (**Khaled traveled sad**)

The Arabic language has the following properties:

## 3.3 Inflected language:

It is a language in which lexical units vary in number and in bending (the number of names, or verb tense) according to the grammatical relationships they have with other lexical units.

The inflected system displays a varied marking. For example, Arabic has three nominal cases: nominative (NOM), which is the default case, the accusative (ACC) for verbal complements and the genitive (GEN) for the dependent of a preposition.[20]

```
كتب الأولادُ        kataba AlAwlAd+u
                    (V)PAST (N)+NOM
                    Wrote  kids
                    "The kids wrote"

قابل سمير الأولادَ   qAbAlA samir AlAwlAd+a
                    (V)PAST (N) (N)+ACC
                    met samir childrens
                    "samir met childrens"
```

سلّم سمير على الأولادِ salama samir 3ly AlAwlAd+i
(V)PAST (N) (PREP) (N)+GEN
welcomed samir children
"samir welcomed children"

Note that the mark of the definite is made by the set of letters (al-) while the mark of the indefinite is made by a diacritic fused with sign of the short vowel, the sign is called "tanwiin" (an, an, in ):

**al**+Awlaad +**u**     **vs**     Awlaad+u+**n**
**DEF**+(N)+NOM           (N)+NOM+**INDEF**
**The child**              **a child**

### 3.4 Voyellation

An Arabic lexical unit is written with consonants and vowels. Vowels are added above or below letters. They are required for reading and for a correct inderstanding of a text and they can differentiate lexical units having the same representation.

Example: the word non vowelized "رسل" (RCL) can have 13 different voyellations:

| Voyellation | Transliteration | Traduction | grammatical category |
|---|---|---|---|
| رَسَلَ | Raçala | Has been long and flowing | Verb, accomplished, active voice, 3rd person, masculine singular |
| رَسَّلَ | Rassala | he read with slowdown | Verb, accomplished, active voice, 3rd person, masculine singular |
| رُسِّلَ | Russila | Has been read with slow-down | Verb, accomplished, active voice, 3rd person, masculine singular |
| رَسِّلْ | Rassil | Do read slow-down | Verb, Imperative, 2nd Person, masculine, singular |
| رُسُلُ | Rusulu | Prophets | Noun, masculine, plural, nominative determined |
| رُسُلَ | Rusula | | Noun, masculine, plural,, accusative determined |
| رُسُلِ | Rusuli | | Noun, masculine, plural, genitive determined |
| رُسُلٌ | Rusulun | | Noun, masculine, plural, nominative indetermined |
| رُسُلٍ | Rusulin | | Noun, masculine, plural, genitive indetermined |
| رِسْلٌ | Rislun | tender | Noun, masculine, singular, nominative indetermined |
| رِسْلٍ | Rislin | | Noun, masculine, singular, genitive, indetermined |
| رَسْلٌ | Raslun | | Noun, masculine, singularr, nominative, indetermined |
| رَسْلٍ | Raslin | | Noun, masculine, singulier, genitive, indetermined |

**Table1 :** different voyellation of « رسل »

### 3.5 The agglutination

The Arabic language is agglutinative that clitics stick to nouns, verbs, adjectives which they relate. These phenomena pose formidable problems for the automatic analysis of Arabic, as in so far as that they greatly increase the rate of ambiguity by introducing additional ambiguities in the segmentation of words. Indeed, an Arabic word may have several possible divisions: proclitic, flexive form and enclitic.

❖ **The agglutination for verbs :**
As shown in the table below, the maximum possible for chaining a verbal base is equal to five: an inflected form with four other morphemes representing the clitic that are attached.[20]

| Proclitics | | | Base | Enclitic |
|---|---|---|---|---|
| Article interrogation | Conjonctions | Markers / particles | verb | personals pronouns |
| The article of interrogation "أ" | Coordinating conjonctions: "و" et "ف" | particle subjunctive "لِ":"نصب" (for) | Inflected form | 1st person |
| | | The marker of corroboration "لَ":« تأكيد » | | 2nd Person |
| | | The particle of the future: « س » | | 2nd Person |

**Table2:** the constituents of a verbal form agglutinated.[20]

But there are several grammatical restrictions that apply to manage the possible combos of a verbal base with different clitics.
- Article of interrogation cannot be stuck with a verb in the imperative or subjunctive.
- The proclitic "لِ" is considered as a subjunctive particle, it do not stick with a flexed form in the present tense.

- The particle of the future "س" cannot be adjoined to a verb conjugated in the Imperfect, active or passive voice.
- Personal pronouns do not stick to verbs in the passive voice or verbs intransitifs.[20]

  ❖ **The agglutination of nouns :**
  The possible combos are presented in table 3 :

  | Proclitics | | | | base | Enclitics |
  |---|---|---|---|---|---|
  | Article of interrogation | Conjonctions | presentations | Defined Article | noun | personals pronouns |
  | The article of interrogation "أ" | Coordinating conjunctions: "و" et "ف" | بِ (bi) لِ(li)كَ(ka) | ال | Inflected form | 1st Person<br>2nd Person<br>3rd Person |

  **Table 3**: Components of a verbal form agglutinated.[20]

The maximum number of concatenation of names is also five: an inflected form with four morphemes (four proclitic or three proclitic and an enclitic).
Agglutination rules are summarized as follows:
- The definite article "ال" cannot coexist with a personal pronoun.
- The definite article "ال" cannot coexist with an inflection undefined, especially the "tanwin".
- A preposition as ب (bi), ك (ka) or ل (li) can co-exist with an inflection at genitive.[20]

## 3.6 Pro-drop (= to an empty pronominal subject)

The ASM neglects systematically the morphological realization of subject pronoun. However, the verb agrees in person, number and gender with the pronoun omitted, as the following example shows: (The pronoun call is placed between brackets).

    *akaluu {humu}*     *vs*     *akalnna {hunna}*
    (V)PAST.3.MASC.PL            (V)PAST.3.FEM.PL
    have eaten {they}              have eaten {they}
    'they have eaten.'(أكلوا)       'they have eaten.'(أكلن) [21]

## 3.7 The grammar

Traditional grammar has characterized by:

### 3.7.1 The morphology
Which is divided into:
- **a. Derivational morphology (الاشتقاق)**, studying the construction of words and their transformation according to the desired meaning.
  The main branches are:
  ✓ The verbal noun: Each verb has at least a verbal noun which expresses the same semantics of the verb. as the verb "وَدَّ" (Wadda) that admits as verbal noun "وُدٌّ" (wuddun)

- ✓ The active participle: is a name associated with any action verb which means the agent of the verb that is to say who do the action. As the verb "كتب" which admits as active participle"كاتب" following the pattern " فاعل "
- ✓ The passive participle: is a name associated with any verb which means the agent who is undergoing the action. Example: "كَسَرَ" is a verb that admits the passive name "مكسور" following the pattern "مفعول"
- ✓ The name of the place (or time) is a deverbal which means the place (or time). Example: "درس" ▢ "مدرسة"
- ✓ The name of manner: it indicates how the action expressed by the verbe.[20]

b. **Inflexional morphology** (الاعراب), Arabic is an inflected language concerned by the casual marking for noun and the adjective or the conjugation of verb.

> **Flexion of verbs:**

The conjugation of verb describes the variation of its shape depending on the mode and time used.

- The past "الماضي": it indicates that the action expressed by the verb is completed and it is characterized by the suffix of the verb according to gender, number...
  Example: the verb "ذهب" he adds "ا" for dual masculine and "تا" for female duel.
- The future "المضارع": it indicates that the action expressed by the verb is incomplete and is characterized by prefixation of the verb and one or more suffixation according to gender, number...
- The imperative "الأمر": it expresses the order with the 2nd person singular, dual and plural (or female).

With the exception of the verbs that follow regular forms of conjugation, there are irregular verbs requiring special handling.[20]

> **Flexion of nouns :**

The nominal Arabic system has various systems of declination. We can distinguish:

- *Declination of base (singular):* it takes the vowel "ضمة" as nominative and the vowel "فتحة" as accusative and the vowel "كسرة as genitive. When the name is undefined, the" tanwin "is marked by signs following: (un), (an) and (in).
- *Declination of diptotes (singular):* the example of male adjectives of colors which they follow the scheme "أفعل" as "أصفر" and follow the feminine the scheme "فعلاء" as "بيضاء".
- *Declination of deverbal of defective roots:* as an example of passive participles that end in "ى" or "ا" as "معفى", they lose their inflection. There are also verbal nouns which have a form such as "قاض" and are generally al-

- tered to "قاضي" by adding the glide "ي" at the end of the initial form.
- *Declination of the name in the duel:* To define the dual of a defined or undefined name, we suffix him by "ان" in nominative and "ين" in the accusative and the genitive. But there are always exceptions like the case of words ending by "ء", "ي", "و" whose termination is transformed before adding the suffix of duel. Example: "ملهى" à "ملهيين".
- *Declination of the plural noun:* For regular masculine plural we suffix it by "ون" or "ين" as "مفكر" à "مفكرون". For regular feminine plural we suffix it by "ات" as "وسادة" à "وسادات".

  Of course there is the broken plural nouns that require changes and infixation of their singulars and whose do not follow own variations.

- ➢ **Flexion of tools word:** There are tools words not declinable such as "إلى" and other declinable as the quantifier "كلٌّ" which he accepts the final three vowels of the nominative, accusative and genitive.[20]

### 3.8 Recent classification of Arabic units

To our knowledge, studies that are made to classify lexical units of the Arabic language according to parts of speech are very few[21]. Recent approaches for classification of lexical units fall into two approaches: Some consist of a classification identified for Indo-European languages without taking into account the exceptions of one language or another, note the possible existence of a class in a language and not in the other. Others have retained the traditional classification Arabic then introduced some modifications.

A classification is fairly recent one made by Khoja[22] in the development of a morphosyntactic tagger. Khoja presents a label based on the traditional classification and refined by the subdivisions proposed by Haywood[23]. Under this classification, the lexical units are divided into five classes: noun, verb, particle, residual and punctuation. Some are refined into sub classes illustrated in Figure 4.

**Hyp7:** the type of exercise most used is the text with blanks;

In our questionnaire (see Appendix) we asked a group of teachers (60 teachers, fewer but just to know the opinion of a community of teachers) whose have experience ranging from 4 years and 26 years and whose teach students ranging in age from 7 years and 18 years, that is to say, from primary to secondary level.

According to the harvest results of question 1 of our questionnaire, we obtain the results summarized in Table 4 and Figure 5.

| statistical variables | Effectives (ni) | frequency (fi) |
|---|---|---|
| تكتبه (you write it) | 03 | 05% |
| تبحث عنه (you search it) | 00 | 00% |
| إحدى الحالتين, حسب السياق (one of the two, depending on the context) | 57 | 95% |
| Total | 60 | 100% |

**Table 4**: Answer to question 1 of the questionnaire

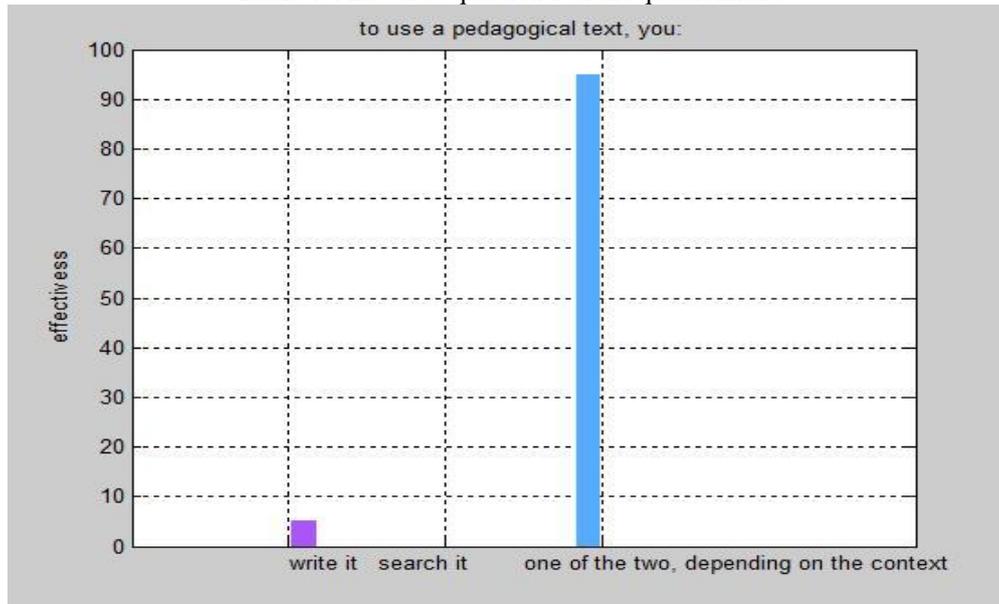

**Figure5:** Diagram showing the results of the 1$^{st}$ question of the questionnaire

In question 1 which he questioned the method of obtaining the text, we found that 5% of teachers write their own texts and 95% of teachers, either they seek it either they write it, and these depending on the context, which proves particularly and not entirely our **Hyp1**, as there is no teacher that searches automatically the text but rather depends on teaching conditions (the objective of the course, student level, ...).

Question Number 2 is composed of three parts:
- In a first step we asked teachers if they have personal collections, which led us to a result which the majority of teachers (95%) responded 'yes'.

- Among the teachers who answered 'yes', we have 92.98% of them who have oganized collections.

These collections are organized as shown in Table 5 by:
- ✓ The subject
- ✓ The students' level
- ✓ The course

| statistical variables | Effectives (ni) | frequency (fi) |
|---|---|---|
| حسب الموضوع (according on the subject) | 03 | 5.66% |
| حسب مستوى التلاميذ (according on the level of students) | 03 | 5.66% |
| حسب هدف الدرس (according to the objective of the course) | 47 | 88.67% |
| Total | 53 | 100% |

**Table 5:** Response to question2 (3$^{rd}$ part) of the questionnaire

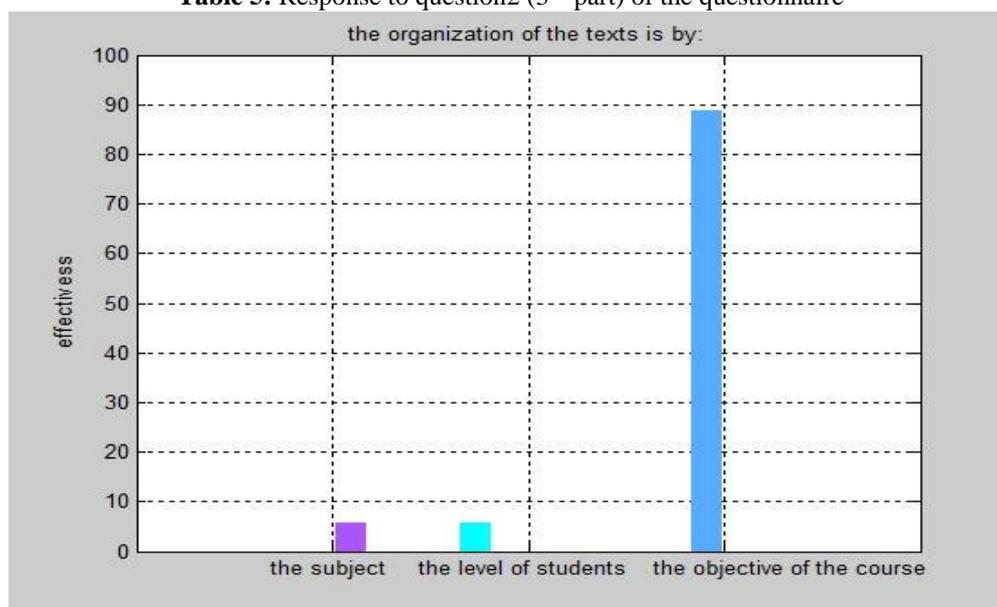

**Figure6:** Diagram showing the results of the 2$^{nd}$ question in the questionnaire for its 3$^{rd}$ party

According to Table5 and Figure6 representative of the table we deduced that most teachers have an organized collection according to the objective of the course and then come the subject and the level of students in 2nd place with equal percentages.

Question 3 deals with the manner in which the teacher obtained his texts by giving three proposals (see table 6 and figure 7):
- Search for a physical activity

- They get it through a personal reading
- Are in the official program

| statistical variables | Effectives (ni) | | | frequency (fi) | | |
|---|---|---|---|---|---|---|
| | 1st choice | 2nd choice | 3rd choice | 1st choice | 2nd choice | 3rd choice |
| بحث من أجل نشاط معين (Search for a particular activity) | 52 | 03 | 02 | 86.66% | 05% | 3.33% |
| تحصلت عليه من خلال قراءة شخصية (you get it through a personal reading) | 01 | 12 | 12 | 1.66% | 20% | 20% |
| موجودة في البرنامج (Exists in the program) | 07 | 43 | 44 | 11.66% | 71.66% | 73.33% |
| No response | 00 | 02 | 02 | 00% | 3.33% | 3.33% |
| Total | 60 | 60 | 60 | 100% | 100% | 100% |

**Table 6:** Answer to Question 3 of the questionnaire

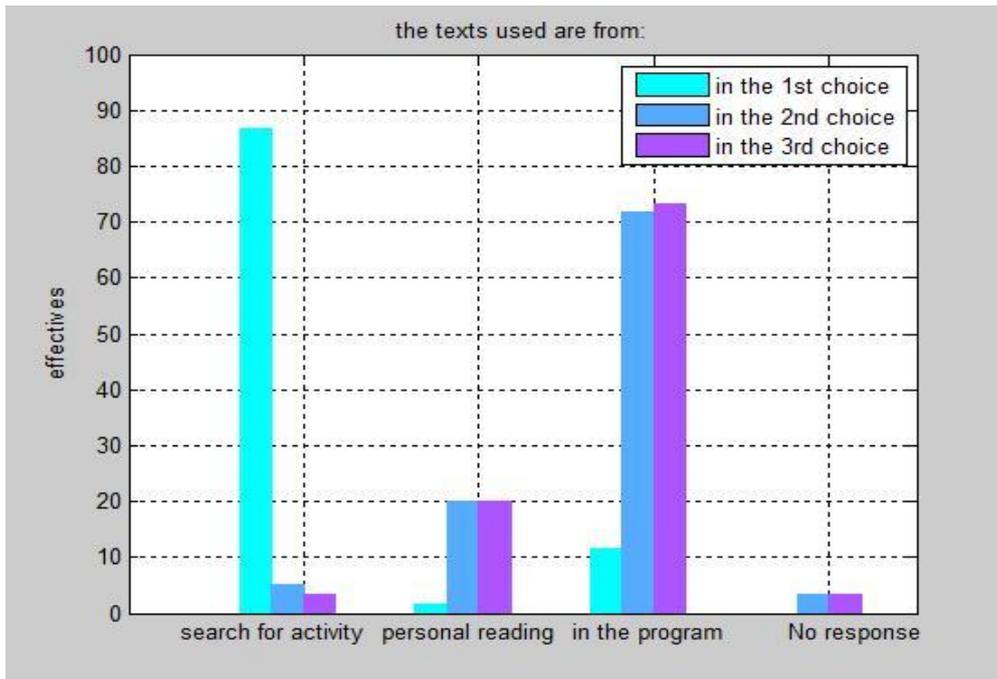

**Figure7:** Diagram showing the results of the 3rd question of the questionnaire

In this question, teachers are required to classify the answers according to priorities. As shown in table 6 and figure7, we set the percentage assigned to each proposal selected by the teacher in the first place, then the percentages of the proposals selected in the 2nd place and finally those who are chosen in 3rd place.

We noticed in this question that there are teachers who have not put in those proposals theirs 2$^{nd}$ and 3$^{rd}$ choices (shown in table 6 by the line called 'no answer').

Analysis of the results led us to conclude that the majority of teachers (86.66%) have obtained texts when searching for a specific activity, then come in 2nd place predefined text in the program with a percentage of 71.66%.

This finding allowed us to prove the **Hyp2** and thereafter we will be based on the text searched for a specific activity in the first place, then the texts of the program.

When we had to prove or refute the **Hyp3**, we focused on question 6 which these results are summarized in table 7 and illustrated by Figure 8.

| statistical variables | Effectives (ni) | frequency (fi) |
|---|---|---|
| نعم(yes) | 59 | 98.33% |
| لا(no) | 01 | 1.66% |
| **Total** | 60 | 100% |

**Table 7:** Answer to Question 6 of the questionnaire

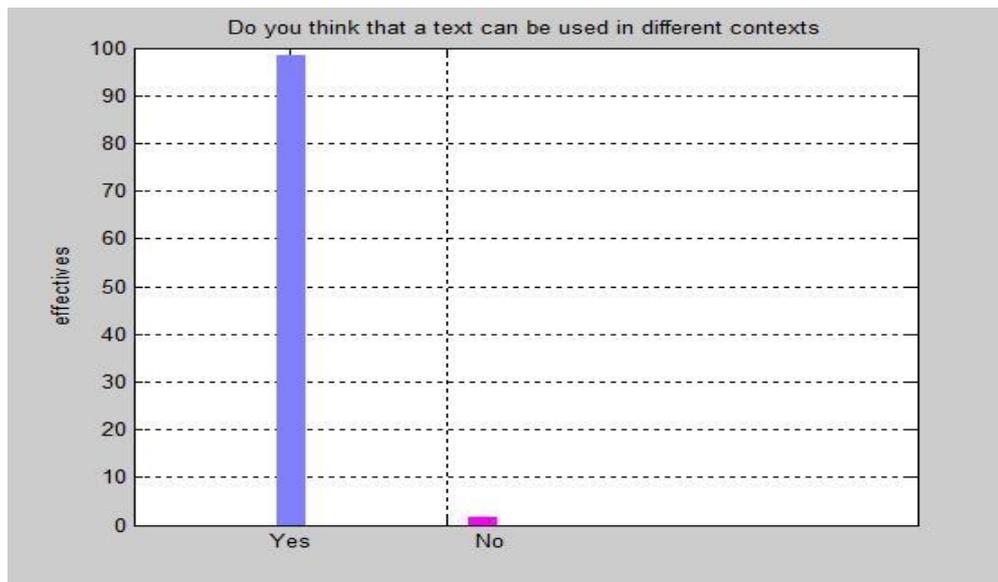

**Figure8:** Diagram showing the results of the 6th issue of the questionnaire

We found that a percentage of 98.33% of the teachers confirmed the **hyp3** concerning the use of a text in different contexts, and 1.66% (a single teacher) which held that each text has its own context. This allows us to prove our hypothesis.

Question number 7 asks the teacher about what criteria the most insists on his choice of text, with giving him three proposals (see Table 8 and Figure 9):
- Certain statements in the context of the lesson
- The subject on which the text unfolds
- the length of the text

Developing a model for a text database indexed pedagogically for teaching the Arabic language

| statistical variables | Effectives (ni) | | | frequency (fi) | | |
|---|---|---|---|---|---|---|
| | 1st choice | 2nd choice | 3rd choice | 1st choice | 2nd choice | 3rd choice |
| عبارات معينة في سياق الدرس (Certain statements in the context of the lesson) (1) | 49 | 08 | 00 | 81.66 | 13.33% | 00% |
| الموضوع الذي يدور حوله النص (The subject on which the text unfolds) (2) | 08 | 18 | 07 | 13.33% | 30% | 11.66% |
| طول النص (the length of the text) (3) | 00 | 07 | 23 | 00% | 11.66% | 38.33% |
| (1)+(3) | 03 | 00 | 00 | 05% | 00% | 00% |
| (2)+(3) | 00 | 25 | 00 | 00% | 41.66% | 00% |
| No response | 00 | 02 | 30 | 00% | 3.33% | 50% |
| **Total** | 60 | 60 | 60 | 100% | 100% | 100% |

**Table 8:** Answer to Question 7 of the questionnaire

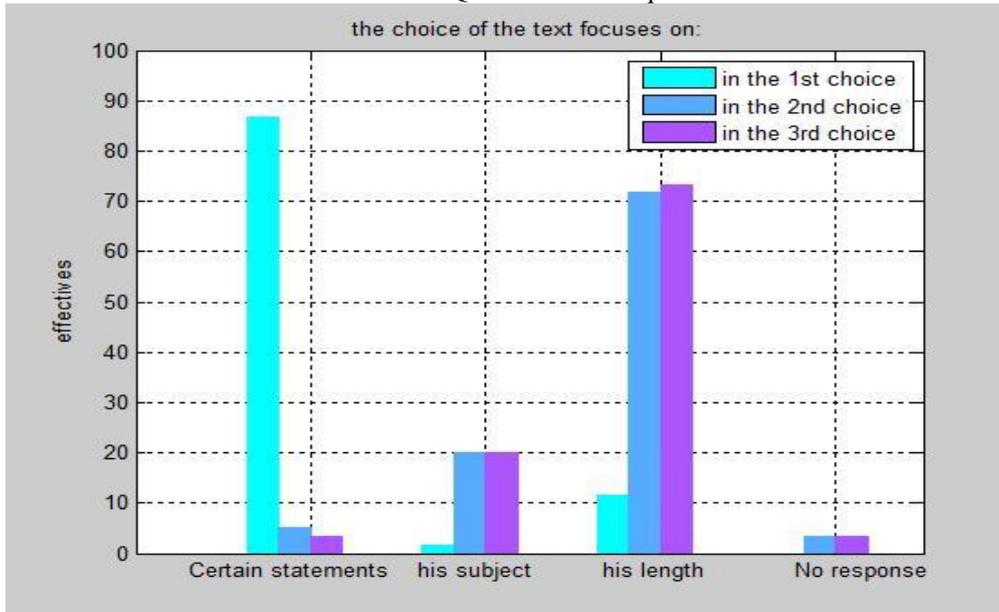

**Figure9:** Diagram showing the results of the 7th question of the questionnaire

From Table8 and Figure 9 we see that teachers focus in the first rank on certain statements in the context of the lesson (81.66%) and in the 2nd place on the subject and the text length. This proves the Hyp4.
On the other hand the length of the text has an effect on his choice, indeed all teachers confirm the **Hyp5** with a percentage of 100%. We also deduced that all the teachers are always try to choose a length of text most minimum possible (see table 9).

Asma Boudhief, Mohsen Maraoui, Mounir Zrigui

| statistical variables | Effectives (ni) | frequency (fi) |
|---|---|---|
| الأكثر طولا (the longest) | 00 | 00% |
| الأقلّ طولا (the shortest) | 60 | 100% |
| **Total** | 60 | 100% |

**Table 9:** answer to the question 8.A of the questionnaire

Regarding the use of text, we deduced that all teachers use texts for inflectional morphology and derivational morphology (Question9), but they are mainly used it for derivational morphology with different categories (Question9.C), which proves the **Hyp6**.

To learn more about what type of inflectional and derivational morphology teachers use the text, we proposed the two issues which Q9.B Q9.A and their results are made in tables 10 and 11 and their presentations in Figures 11 and 12.

| statistical variables | Effectives (ni) | frequency (fi) |
|---|---|---|
| التصريف في أزمنة مختلفة مع ضمير واحد (conjugate the verb in different time with a single pronoun) (1) | 00 | 00% |
| التصريف في زمان معين مع ضمائر مختلفة (conjugate the verb in a specific time with different pronouns) (2) | 00 | 00% |
| التصريف في أزمنة مختلفة مع ضمائر مختلفة (conjugate the verb in different time with different pronouns) (3) | 51 | 85% |
| (2)+(3) | 02 | 3.33% |
| (1) +(2) +(3) | 07 | 11.66% |
| **Total** | 60 | 100% |

**Table 10:** Answer to the question 9.A of the questionnaire

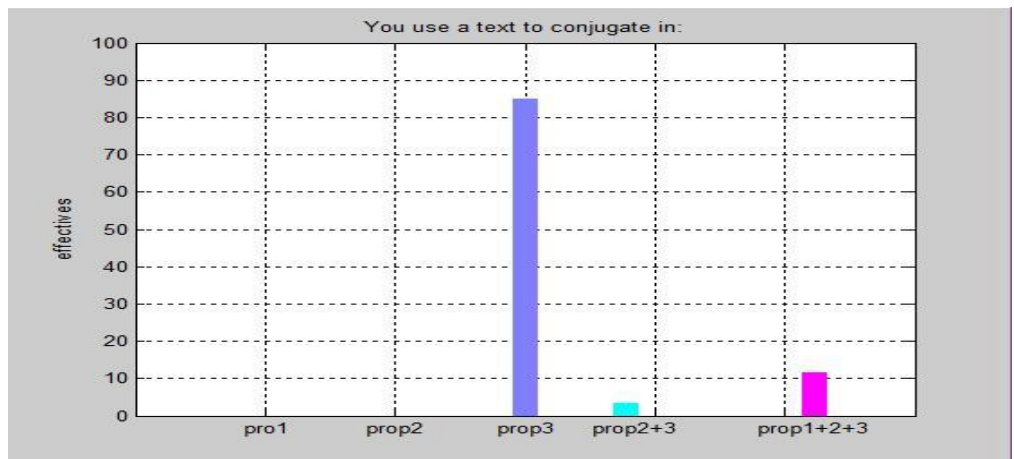

**Figure11:** Diagram showing the results of the questionnaire the question 9.A

| statistical variables | Effectives (ni) | frequency (fi) |
|---|---|---|
| نوع الجمل (types of sentences) (1) | 02 | 3.33% |
| نوع الأسامي (types of nouns) (2) | 00 | 00% |
| نوع التراكيب (types of compositions) (3) | 01 | 1.66% |
| (1)+(3) | 03 | 05% |
| (1) +(2) +(3) | 54 | 90% |
| **Total** | **60** | **100%** |

**Table 11:** Answer to the question of the questionnaire 9.B

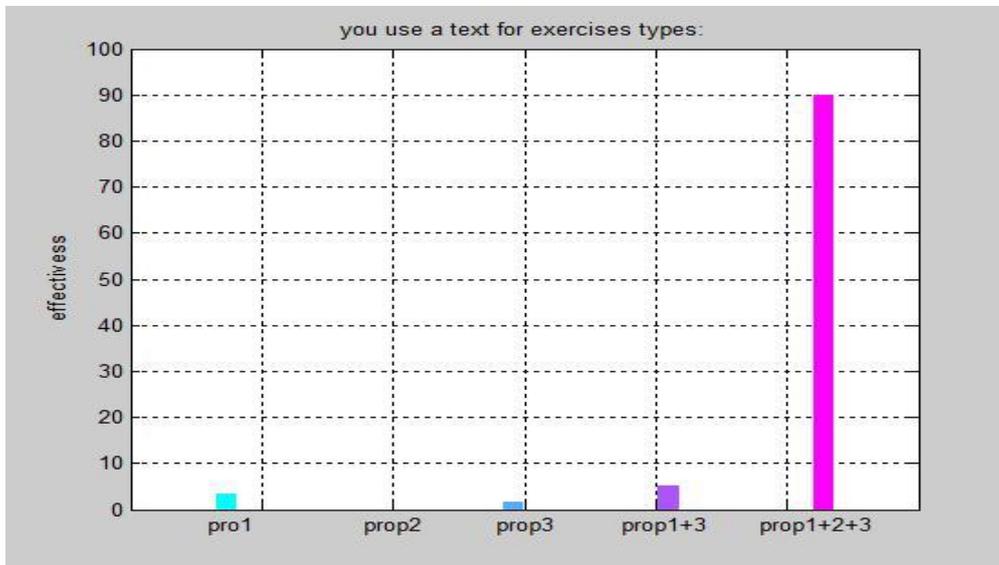

**Figure12:** Diagram showing the results of the question 9.B of the questionnaire

According to previous results, we find that teachers use verbs with varying times and different pronouns, they do not focus on one time or one pronoun. On inflectional morphology, we find that they do not exploit the texts for one type, but they exploit it either for the types of sentences, the types of pronouns or the types of compositions.

All the teachers also judge that the use of texts is made especially to inflectional morphology than derivational morphology.
We know that teachers are using variations of exercises, which lead us to ask a question about the type most used. The results are collected in Table 12 and represented by the Figure13.

Asma Boudhief, Mohsen Maraoui, Mounir Zrigui

| statistical variables | Effectives (ni) | | | | frequency (fi) | | |
|---|---|---|---|---|---|---|---|
| | 1st choice | 2nd choice | 3rd choice | number «0» | 1st choice | 2nd choice | 3rd choice |
| ملئ الفراغات بما يناسب (fill in the blanks) (1) | 52 | 03 | 01 | 02 | 86.66% | 05% | 3.33% |
| أسئلة متعددة الإختيارات (QCM) (2) | 04 | 49 | 01 | 03 | 6.66% | 81.66% | 05% |
| سؤال و جواب (question/Response) (3) | 03 | 04 | 34 | 15 | 05% | 6.66% | 25% |
| (1)+(2) | 01 | 00 | 00 | 00 | 1.66% | 00% | 00% |
| (2)+(3) | 00 | 02 | 00 | 00 | 00% | 3.33% | 00% |
| No de response | 00 | 02 | 24 | 40 | 00% | 3.33% | 66.66% |
| Total | 60 | 60 | 60 | 60 | 100% | 100% | 100% |

**Table 12:** Answer to the question 10 of the questionnaire

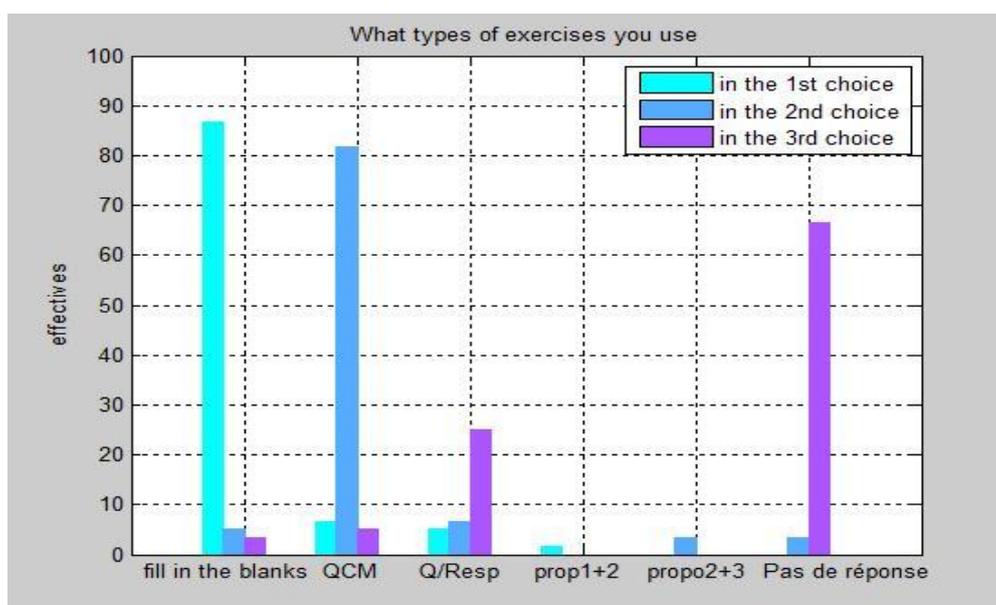

**Figure13:** Diagram showing the results of question 10 of questionnaire

We denote the number "0" the types of exercises that teachers do not use them.
According to statistics, we noticed that teachers (86.66%) use the text in the first place for exercises "fill in the blanks" and 2nd place (81.66%) multiple choice questions "QCM", which proves the **hyp7**.

# 5 A proposed model

## 5.1 Influence of intrinsic and extrinsic properties on the choice of a text

The results of previous surveys have enabled us to introduce the notion of pedagogical context, which is a basic concept to define a form of pedagogical indexation for the teaching of Arabic. Indeed, the text search is the search for an activity, i.e. a search for a given educational context.

## 5.2 Notion of facet-prism

The objective is to take account into the influence of pedagogical context on features of the text. For this we introduce the notion of facet-prism.

### 5.2.1 Prism
A prism is applicable to any feature of text that can be useful for a student looking for a specific exercise or teacher for research or selection of text. This is at the prism that creates the pedagogical value-added.
We consider for example the prism associated with the property "text length", which we call $P_{long}$. The text length can influence the decision of a teacher to use it or not.
According to our survey, most teachers or rather the entire want an effective text but with minimum length, so $P_{long}$ is a prism that allows knowing how many lines a text contains.
To design a prism we should define:
- ❖ The facet: it determines the value of the prism.
- ❖ The text: an input data from which we will extract the information needed to calculate the facet.
- ❖ CP: it is also necessary to calculate the facet.
- ❖ The output value: it characterizes the value of the facet.
- ❖ The processing to be carried to obtain the output data from a given text.

### 5.2 .2 Facet
The facet is the central entity: it represents the concepts on which carries the reasoning of teachers.
There is therefore an essential recursive relationship between prism and facet:
- ➢ A prism without facets is considered as a mechanism, but not to a prism.
- ➢ The facets exist only through the prism: without a mechanism to calculate them, it may be a properties but not a facet.

The Couple prism-facet allowing us to make a coherent indexation takes into account the pedagogical context. The prism is the tool allowing when adding a text at the base, the addition of potential criteria.

Then as shown in Figure 14, each text property we associate a prism, whose value is calculated by a facet.

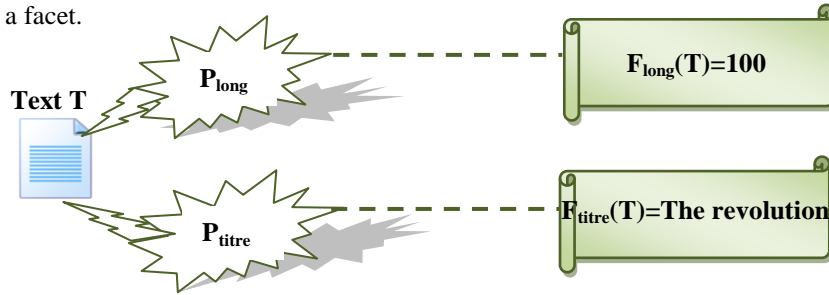

**Figure 14:** Examples of prisms and values for the corresponding facets

## 5.3 The search for text in an indexed text database

The primary function of a text database indexed for language teaching is to allow the search of text.

In our system, we defined the modalities of interaction between such a system and a user-student, or such a system and a user-teacher looking for a text that meets the needs of each with using the concept of facet-prism.

### 5.3.1 The research of activity for a user student

A student must have inscription so that he could authenticated to our system. When the student passed the authentication phase (Figure 15) a new interface appears, representing the categories of exercises offered to him. (Figures 16 and 17)

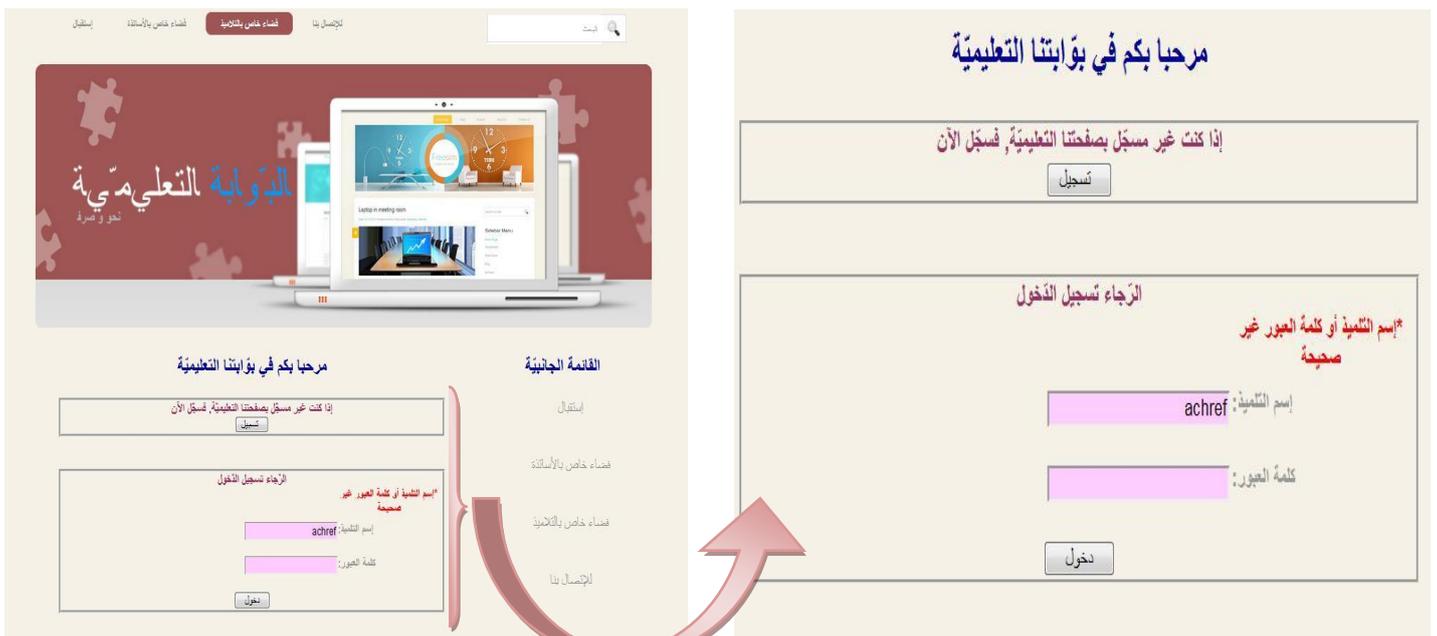

**Figure 15:** interface representing the home page for a user student

Developing a model for a text database indexed pedagogically for teaching the Arabic language

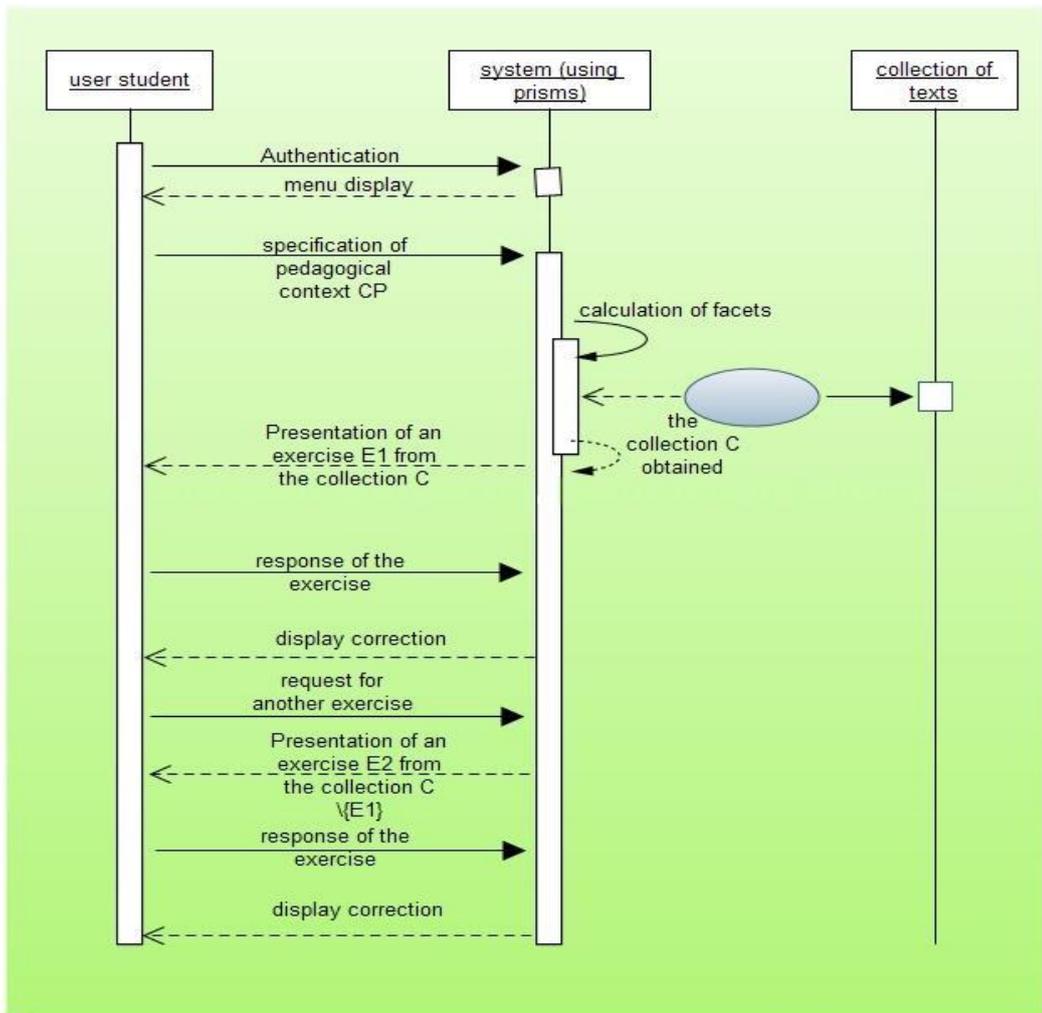

**Figure 16:** Sequence diagram for the research of activity to a user-student

After his authentication, the student is faced with an interface representing the types of exercises available on:
• Morphology (الصرف): conjugation with past, future... with verbs of different classes
• The sentence composition and role of their components

The exercises are characterized by levels of difficulty depending on the complexity of the composition of the sentence and the complexity of verb conjugation. (Figure 17)

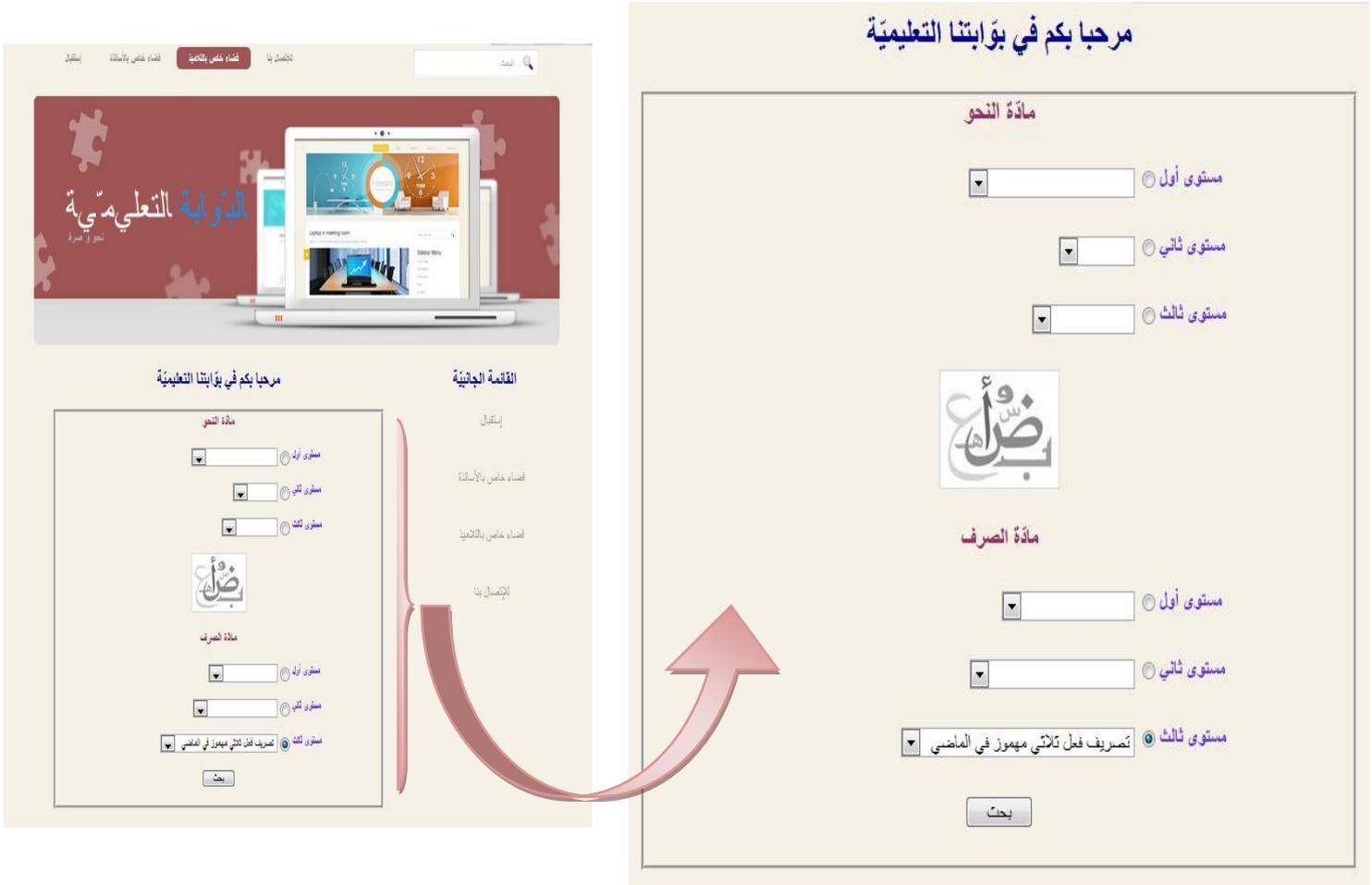

**Figure17:** interface representing the menu displayed for a student

Once a student specifies the pedagogical context (CP), the system performs its processing:
- It calculates the necessary facets
- It extracts using facets computed before, the collection C corresponding to CP

In the following step:
- The system offers the student an exercise E1 of the collection C
- The student responds to exercise
- The system performs the comparison process between the student's response and the response existing in the database and then displays the correction, by characterizing the wrong answers with red color and the correct answers by the color green.

If the student wants to try another exercise in the same class, he asks the system, the latter displays a new exercise E2∈C\ {E1}, and so on until the student's request for stop the process.

Developing a model for a text database indexed pedagogically for teaching the Arabic language

Our system contains a variety of types of exercises:
- ➢ Text hole: in this type of exercise, the system gives either the set of words to fill (Figure 18), either in each empty zone, the system provided a set of proposal as a "text select" (Figure 19), these proposals are selected depending on the type of the correct answer, i.e. if we have a conjunctive preposition type response, the system gives the learner a list of prepositions of this type, which makes our system more dynamic.
- ➢ Multiple choices (Figure 20)
- ➢ Question / Answer (Figure 21)
- ➢ ... Etc.

**Figure18:** Interface representing an exercise with type text blanks

**Figure19:** Interface representing another exercise with type text blanks

**Figure20:** Interface representing an exercise with type multiple choice question

**Figure21:** Interface representing an exercise with type question / response

### 5.3.2 The research of text for a teacher

A user teacher submitted into the space reserved for teachers must:
- ➢ Either register
- ➢ Either fills in the login and password so it can reach other interfaces specific to teachers.

In the case where a teacher is already registered and it has captured the username and password, it will redirect to the rechEnseignant.php (Figure 22) in which he has two possibilities:
- ➢ Either add a new text in our corpus
- ➢ Either the search for an existing text in our corpus, specifying the desired characteristics.

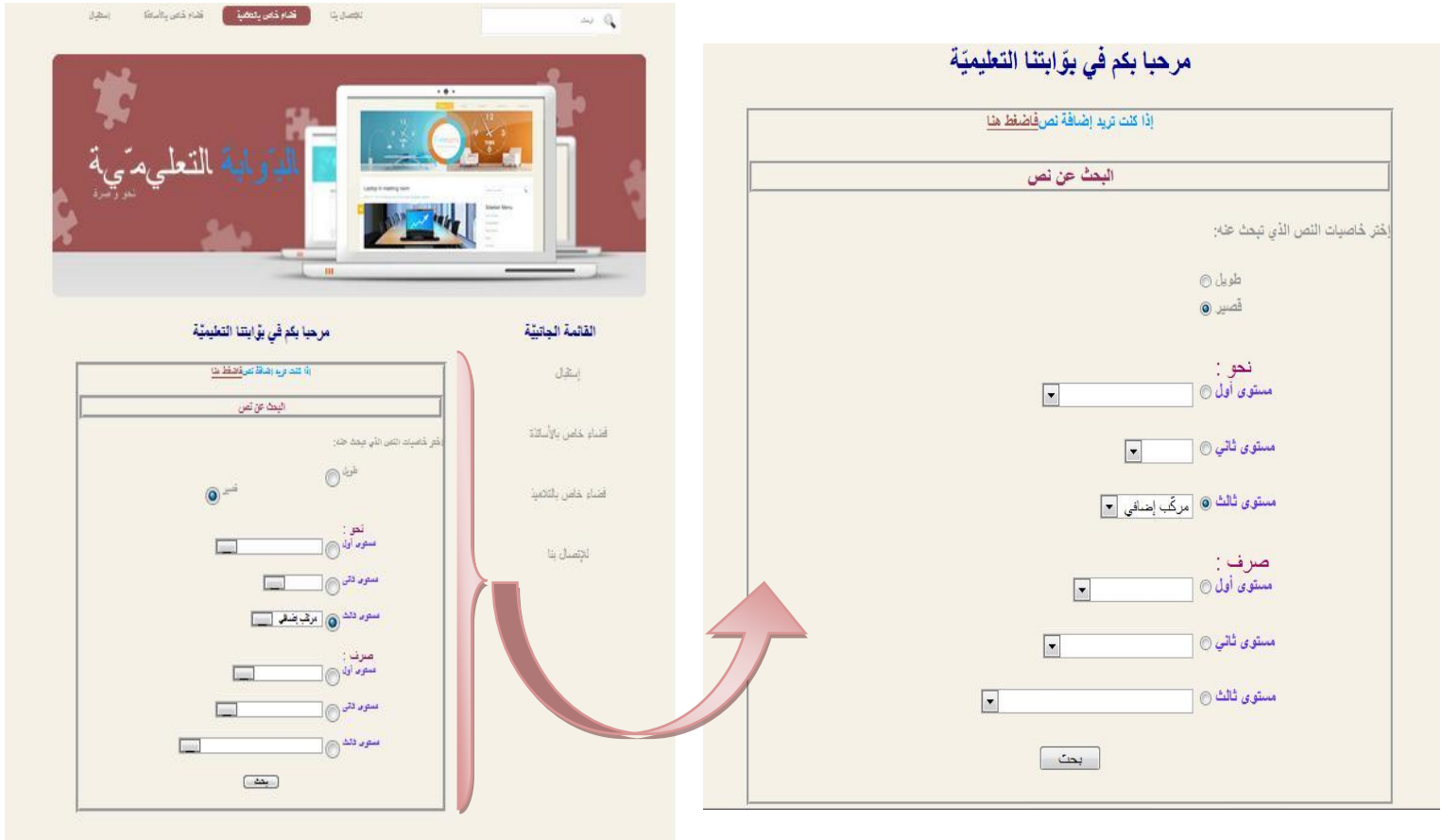

**Figure22:** interface representing the search of text a by teacher

In the case where a teacher wants to search a text for a specific activity, it must specify the intended pedagogical context (CP), ie category and level of complexity of exercises. In that time (Figure23):
- The system calculates the values of facets related to CP
- Seeking the collection C in the base of text
- Presents the collection C to the teacher

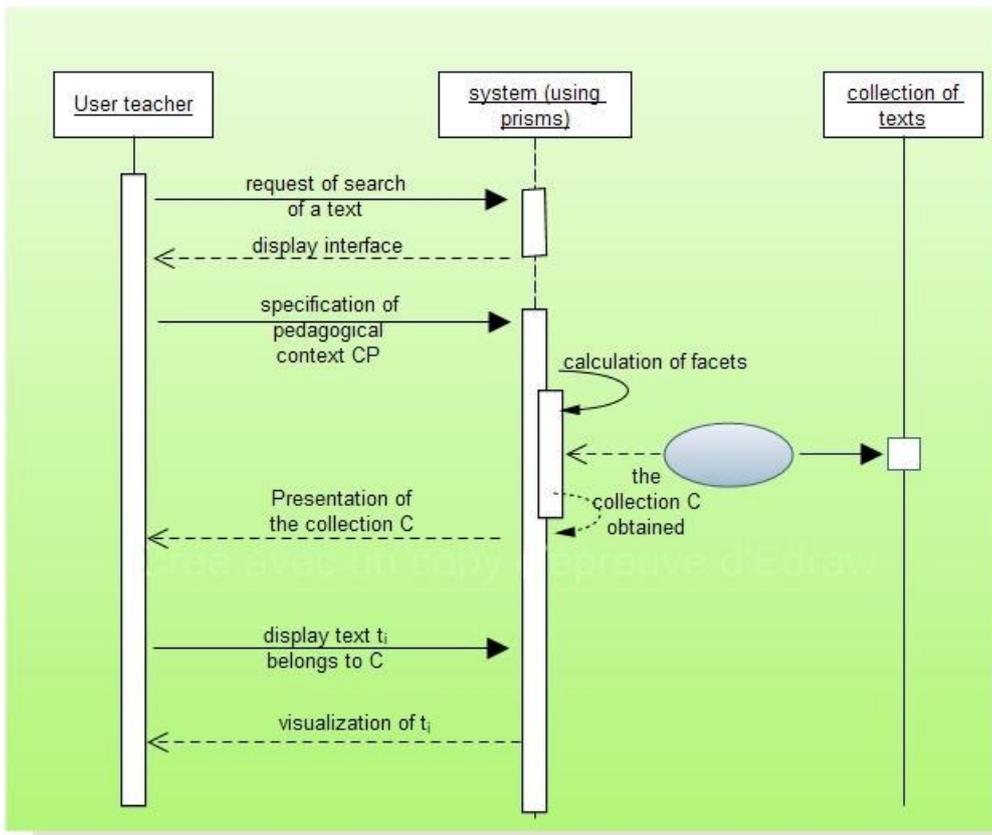

**Figure 23:** Sequence diagram for searching text by a user teacher

As soon as the collection C is ready a new interface is displayed containing a list of texts that meet the requirements of the teacher with a drop-down list containing the types of exercises that a teacher can perform on selected text.

Right now, teachers are asked to determine the number of the selected text ti with the script he wants to apply. (Figure 24)

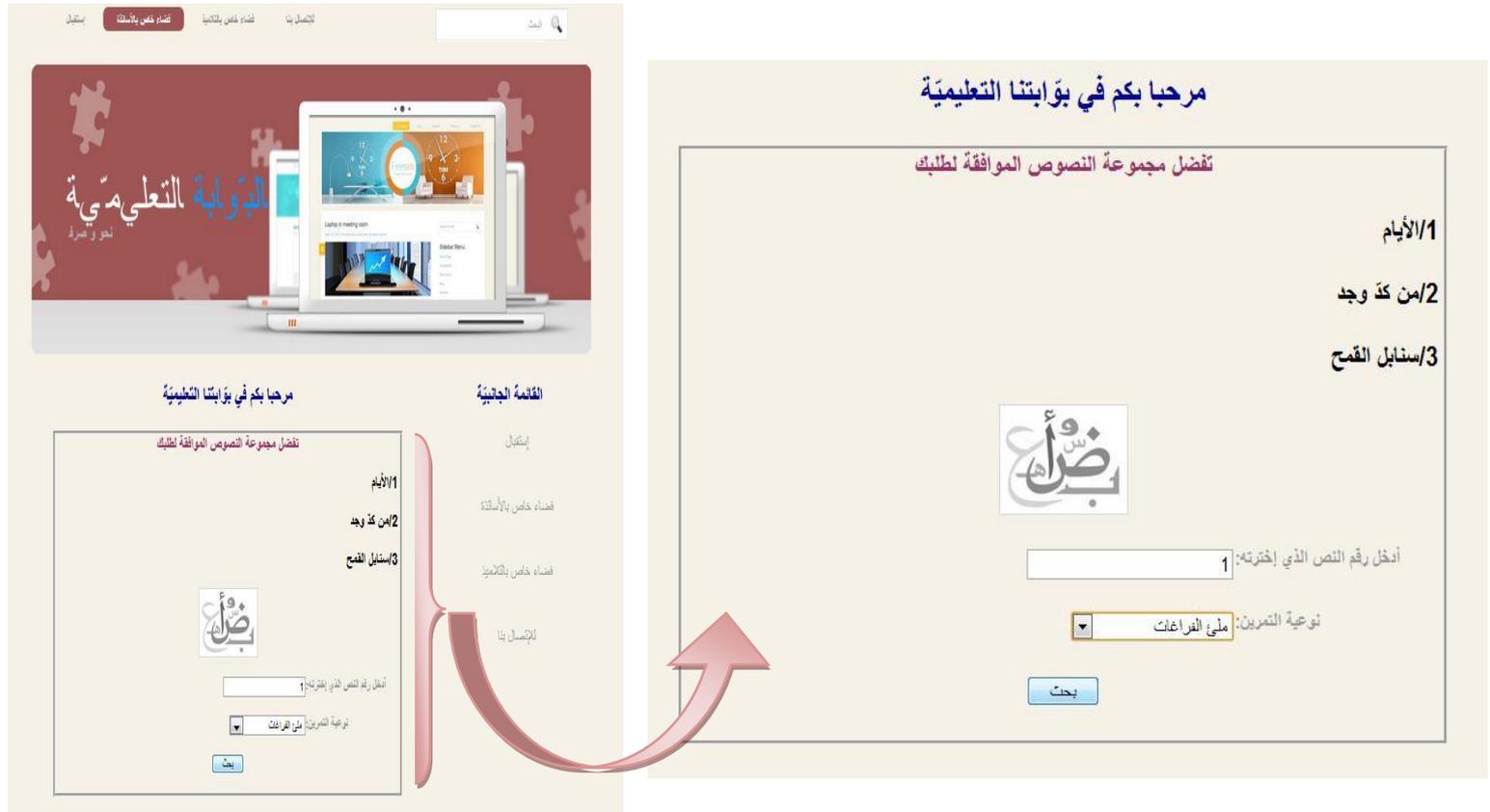

**Figure24:** interface representing the choice of a text and the type of exercise by a teacher

## 5.4 Add of text to the system

All text searched by a student or a teacher is added previously by a teacher as shown in Figure 25.

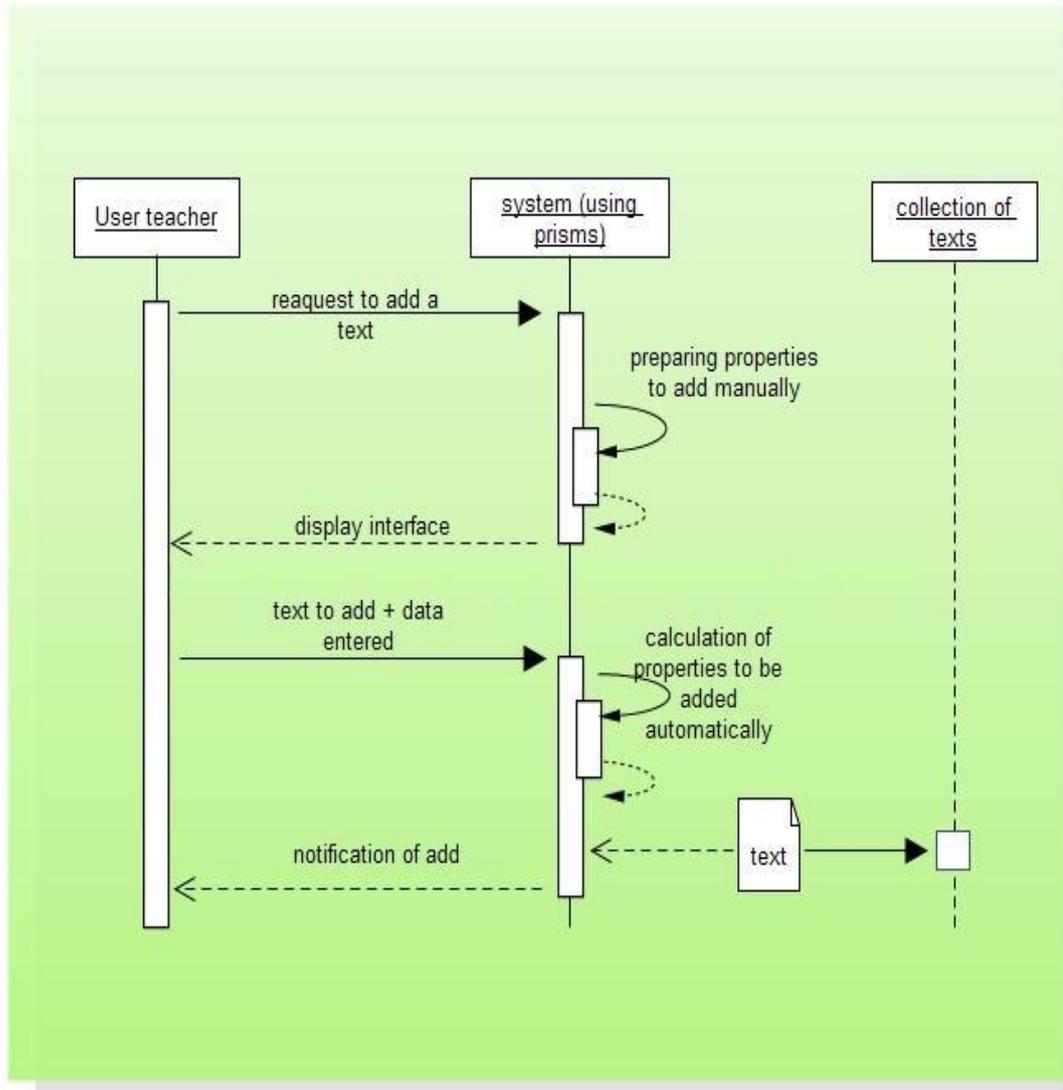

**Figure 25:** Sequence diagram for adding a text by the teacher

As part of a use's sequence, a teacher wanting to add a text to a system start by sending him a request of adds. The system response must identify properties that must be added manually and generate an interface allowing the user to enter the desired information (Figure 26).

Developing a model for a text database indexed pedagogically for teaching the Arabic language

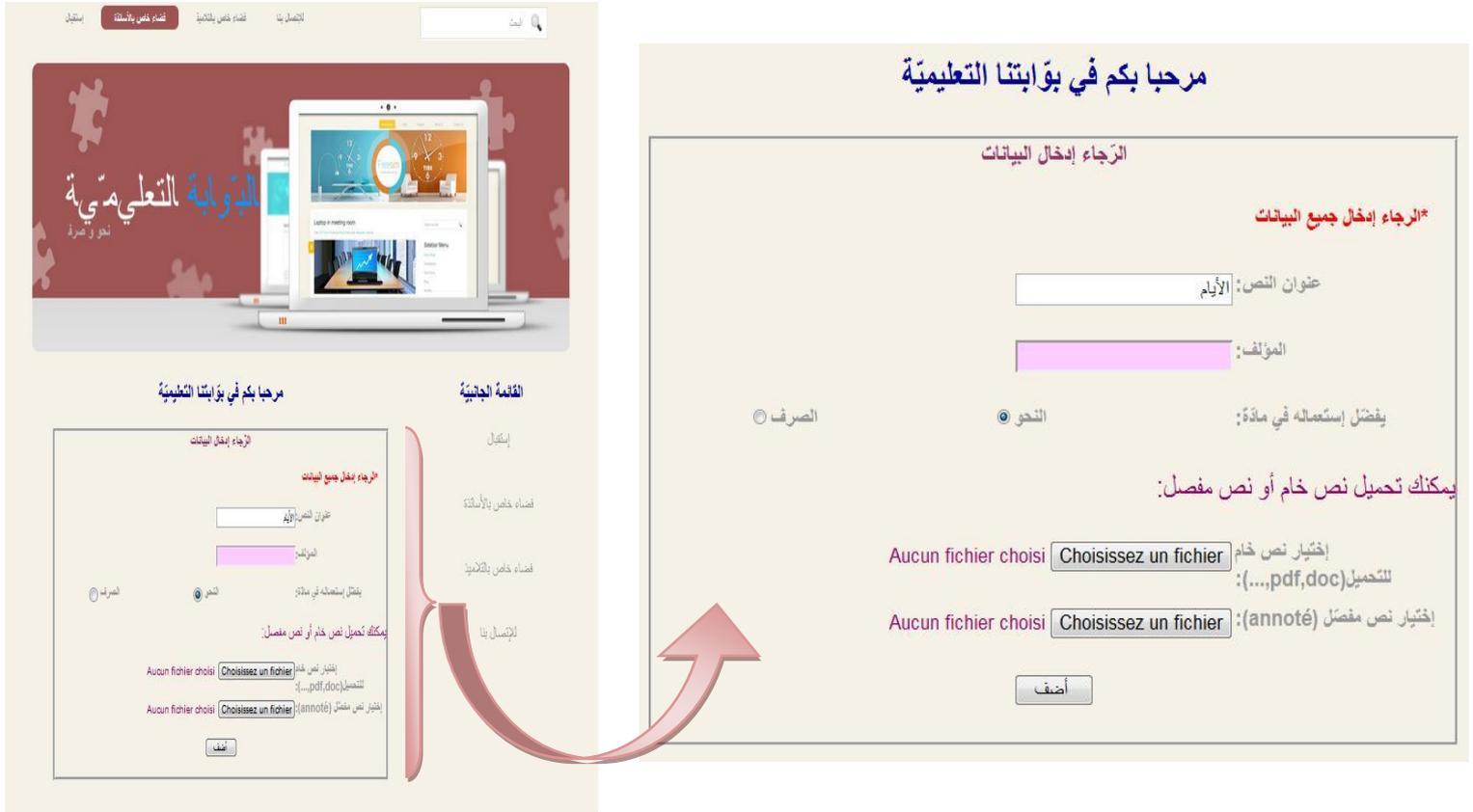

**Figure26:** interface representing the addition of a text by a user teacher

The exchange of teacher with the system will be composed of a text to add and a set of information, some of which will be used as values of facets "constants" as the author or title of text.

The text can be enclosed in two forms:
 ➢ Either a brute text, PDF format or doc
 ➢ Either an annotated text when the system verifies that all information is correctly captured, and the text is an annotated text, it will be automatically stocked in the corpus.

On the other hand, if the system detects that the text is brute; right now you have two choices:
 ➢ Whether the use of a morpho-syntactic analyzer, which will analyze and annotate the text.
 ➢ Either we analyze it a manually

### 5.5 .3 summary diagram of the system

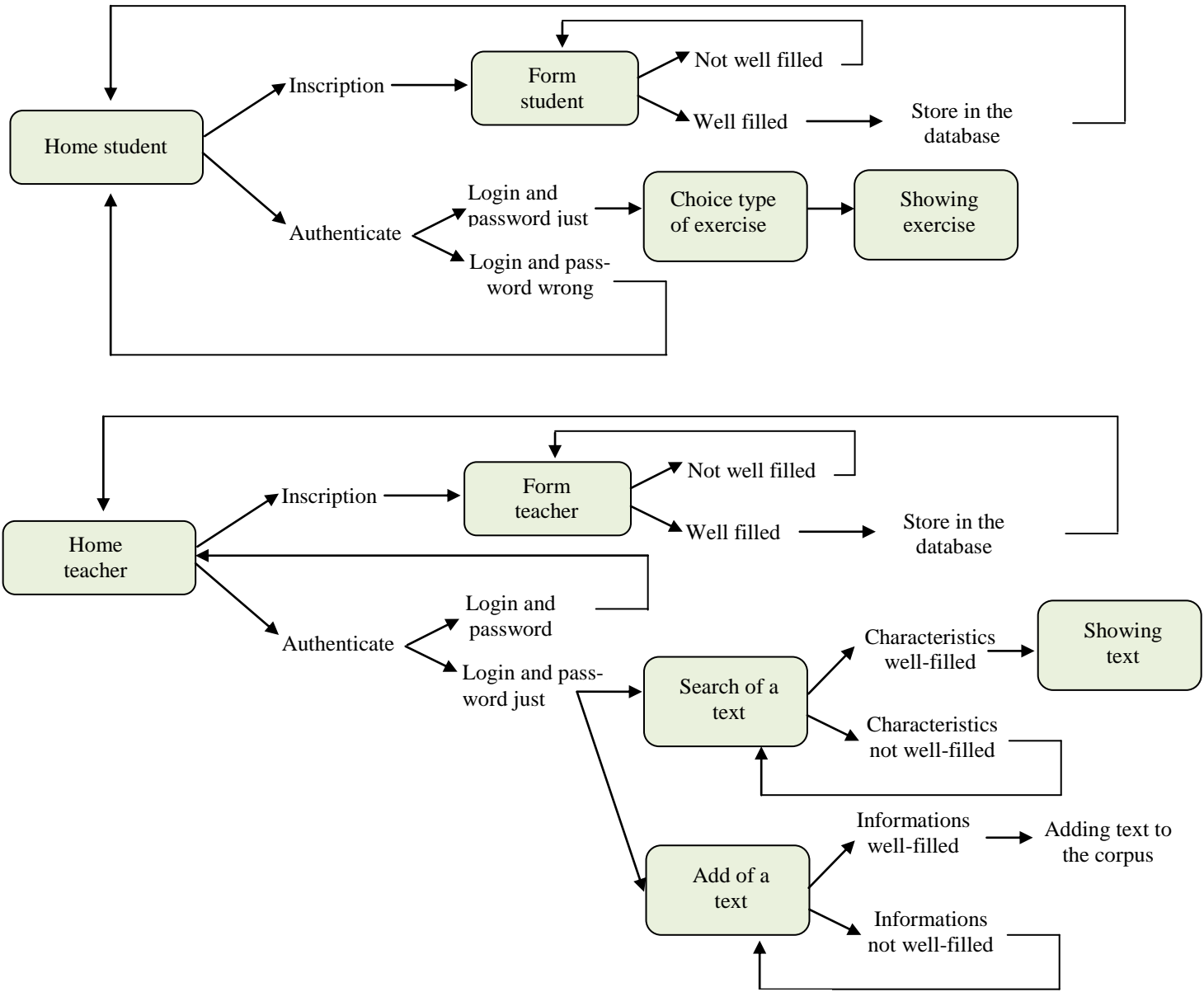

**Figure 27:** general pattern of working of the system

**Conclusion:** As we said at the beginning, we tried in our system outperform existing systems (static) and have a dynamic system that meets the needs of teachers and learners of Arabic.

For a student, the system represents to him an exercise taken from the collection obtained after determination of CP, and if the student asks for more than an exercise in the same category, the system can provide to him and that by choosing each time a new text from collection obtained during the phase of the research.

For teachers, the system represents to him a text's list respondents to their needs with a set of scripts, which he must choose one to apply to the selected text.

So our system offers the possibility of varied types of exercises and texts applied in these exercises with taking into account the educational context and specificity of the Arabic language.

# References (last accessed 2012-03-09)

Asma Boudhief, Mohsen Maraoui, Mounir Zrigui

# Appendix

## إستبيان

**التدريس:**
1) عدد سنوات تدريس اللغة العربية:...............................................
2) إطار التعليم(إعدادي,ثانوي,جامعي,مؤسسة خاصة):...............................
3) العمر و المستوى التعليمي للأشخاص الذين تدرسهم:..............................
..............................................................................

**إستعمال النصوص في إطار التدريس:**

1) عندما تستعمل نص بيداغوجي هل:

☐ تكتبه

☐ تبحث عنه

☐ إحدى الحالتين, حسب السياق

2) هل لديك مجموعة من النصوص الشخصية:

☐ نعم

☐ لا

أ- إذا كانت الإجابة نعم , فهل هذه المجموعة منظمة:

☐ نعم

☐ لا

ب- إذا كانت الإجابة نعم , فكيف(معيار تنظيم النصوص : حسب الموضوع , مستوى التلاميذ , .....):
..............................................................................

3) النصوص التي تستعملها في التدريس متأتية من: (رتب الإجابات حسب الأولوية)

☐ بحث من أجل نشاط معين

☐ تحصلت عليها من خلال قراءة شخصية

☐ موجودة في البرنامج

4) عند البحث عن نص لتدريس , تبحث عنه في : (رتب الإجابات حسب الأهمية)
- ☐ في المجلات/الصحف
- ☐ في كتب أدبية
- ☐ في الأنترنات
- ☐ إجابات أخرى

..................................................................................................

5) هل وجدت صدفة نصّا و قررت الإحتفاظ به للتدريس؟
- ☐ نعم
- ☐ لا

أ-إذا نعم , هل تحتفظ به:
- ☐ لتمرين معين
- ☐ لمستوى إستعاب معين للتلاميذ(فهم سريع , فهم بطيء)
- ☐ في إنتظار إيجاد طريقة مناسبة لإستعماله

6) هل تعتقد أن النص يمكن أن يستعمل في سياقات مختلفة؟(مستويات تعليم مختلفة,تمارين مختلفة,...)
- ☐ نعم
- ☐ لا

أ-هل حصل لك هذا سابقا؟(أن تستعمل نصّا في سياقات مختلفة):
- ☐ نعم
- ☐ لا

7) عند إختيارك للنص هل تركز على: (رتب حسب الأولوية)
- ☐ عبارات معينة في سياق الدرس
- ☐ الموضوع الذي يدور حوله النص
- ☐ طول النص

8) هل أن طول النص يؤثر على إختيارك له؟
- ☐ نعم
- ☐ لا

أ-إذا كانت الإجابة نعم , فهل تختار نصا:
- ☐ أكثر طولا
- ☐ أقل طولا

9)في أي نوع من التمارين تستعمل النص:

- ☐ صرف
- ☐ نحو
- ☐ كلتا النوعين

أ-إذا كنت تستعمل النص في "الصرف" فهل تستعمله لتمارين حول: (يمكن إختيار أكثر من إجابة)

- ☐ التصريف في أزمنة مختلفة
- ☐ التصريف في زمان معين مع ضمائر مختلفة
- ☐ التتصريف في أزمنة مختلفة مع ضمائر مختلفة

ب- إذا كنت تستعمل النص في "النحو" فهل تستعمله لتمارين حول: (يمكن إختيار أكثر من إجابة)

- ☐ نوع الجمل(إسمية/فعلية)
- ☐ نوع الأسامي(مبتدأ,إسم فاعل,...)
- ☐ نوع التراكيب(نعتي,إضافي,...)

ج-إذا كانت الإجابة في"كلتا النوعين" فأي النوعين تبحث له عن نصوص أكثر من الآخر :

- ☐ صرف
- ☐ نحو

10)ماهي نوعية التمارين التي تستعملها: (رتب حسب الأهمية و إذا كان هناك نوع لا تستعمله فضع عليه رقم صفر )

- ☐ ملئ الفراغات بما يناسب
- ☐ أسئلة متعددة الإختيارات
- ☐ سؤال وجواب